\documentclass[review,12pt]{elsarticle}

\usepackage{amssymb}
\usepackage{setspace} 
\usepackage{geometry}
\usepackage{booktabs}
\usepackage{multirow}
\usepackage{makecell}
\usepackage{longtable}
\usepackage{gensymb}
\usepackage{ragged2e}
\usepackage{caption}

\journal{Elsevier}
\makeatletter
\def\ps@pprintTitle{%
 \let\@oddhead\@empty
 \let\@evenhead\@empty
 \def\@oddfoot{}%
 \let\@evenfoot\@oddfoot}
\makeatother
\begin{document}

\begin{frontmatter}



\title{\LARGE Tabular Machine Learning Methods for Predicting Gas Turbine Emissions}

\author{Rebecca Potts$^a$, Rick Hackney$^b$ and Georgios Leontidis$^{c}$}




\address{$^a$Department of Computing Science, University of Aberdeen, Aberdeen, AB24 3UE, UK r.potts.21@abdn.ac.uk}
\address{$^b$Siemens Energy Industrial Turbomachinery Ltd., Lincoln, LN6 3AD, UK richard.hackney@siemens-energy.com}
\address{$^c$Interdisciplinary Centre for Data and AI, University of Aberdeen, Aberdeen, AB24 3FX, UK georgios.leontidis@abdn.ac.uk}


\begin{abstract}

Predicting emissions for gas turbines is critical for monitoring harmful pollutants being released into the atmosphere. In this study, we evaluate the performance of machine learning models for predicting emissions for gas turbines. We compare an existing predictive emissions model, a first principles-based Chemical Kinetics model \cite{hackney2016predictive}, against two machine learning models we developed based on SAINT \cite{somepalli2022saint} and XGBoost \cite{chen2016xgboost}, to demonstrate improved predictive performance of nitrogen oxides (NOx) and carbon monoxide (CO) using machine learning techniques. Our analysis utilises a Siemens Energy gas turbine test bed tabular dataset to train and validate the machine learning models. Additionally, we explore the trade-off between incorporating more features to enhance the model complexity, and the resulting presence of increased missing values in the dataset.

\end{abstract}




\end{frontmatter}


\section{Introduction}

Gas turbines are widely employed in power generation and mechanical drive applications, but their use is associated with the production of harmful emissions, including nitrogen oxides (NOx) and carbon monoxide (CO), which pose environmental and health risks. Regulations have been implemented to limit emissions and require monitoring. 

To monitor emissions from gas turbines, a Continuous Emissions Monitoring System (CEMS) is commonly employed, which involves sampling gases and analysing their composition to quantify emissions. While CEMS can accurately measure emissions in real-time, it can lead to a high cost to the process owner, including requiring daily maintenance to avoid drift. As a result, CEMS may not always be properly maintained, leading to inaccurate or unreliable measurements. 

Predictive emissions monitoring system (PEMS) models provide an alternative method of monitoring emissions that is cost-effective and requires minimal maintenance compared to CEMS, while not requiring the large physical space needed for CEMS gas analysis. PEMS is trained on historical data using process parameters such as temperatures and pressures, and uses real-time data to generate estimations for emissions. 

To develop a PEMS model, it is necessary to validate the model's predictive accuracy using data with associated emissions values \cite{potts2023attention}. In our experiments, we used test bed tabular data consisting of tests conducted over a wide range of operating conditions to train our models. Gradient-boosted decision trees (GBDTs) such as XGBoost \cite{chen2016xgboost} and LightGBM \cite{ke2017lightgbm} have demonstrated excellent performance in the tabular domain, and are widely regarded as the standard solution for structured data problems. 

Previous studies comparing neural networks (NNs) and GBDTs for tabular regression have generally found that GBDTs match or outperform NN-based models, particularly when evaluated on datasets not documented in their original papers \cite{shwartz2022tabular}, while some NN-based methods are beginning to outperform GBDTs, such as SAINT \cite{somepalli2022saint}.

We compare the predictive performance of an industry used Chemical Kinetics PEMS model \cite{hackney2016predictive}, serving as the baseline, against two machine learning approaches: SAINT and XGBoost, to determine how improvements can be made in emissions prediction for gas turbines. 

We observe that, on average, XGBoost outperforms both the original Chemical Kinetics model and the deep learning-based SAINT model for predicting both NOx and CO emissions on test bed data for gas turbines.

\section{Background}

\subsection{Gradient-Boosted Decision Trees}

Gradient-boosted decision trees (GBDTs) are popular machine learning algorithms that combine the power of decision trees with the boosting technique, where multiple weak learners are combined in an ensemble to create highly accurate and robust models. GBDTs build decision trees iteratively, correcting errors of the previous trees in each iteration. Gradient boosting is used to combine the predictions of all the decision trees, with each tree's contribution weighted according to its accuracy. The final prediction is made by aggregating the predictions of all the decision trees. 

\begin{figure}[!t]
    \centering
    \includegraphics[width=\linewidth]{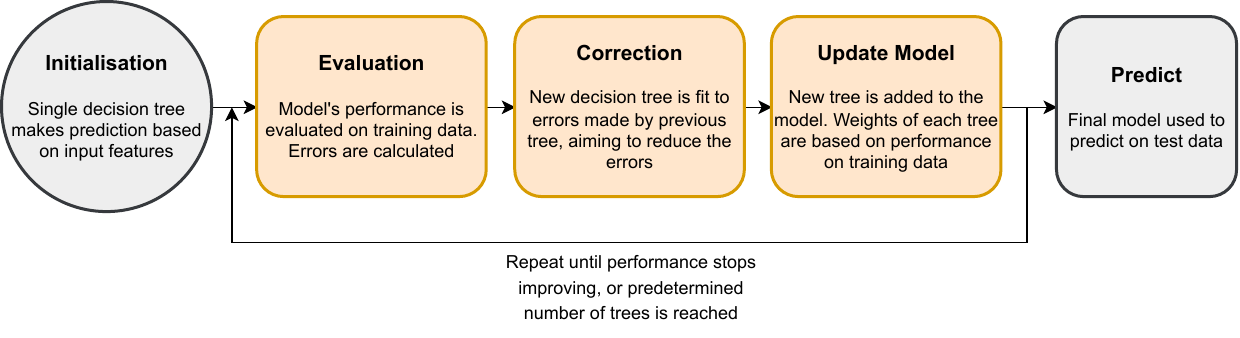}
    \caption{XGBoost initialisation, training, and prediction process.}
    \label{fig:xgboost}
\end{figure}

XGBoost, or eXtreme Gradient Boosting \cite{chen2016xgboost}, is a widely-used implementation of GBDTs, used for both classification and regression tasks. XGBoost is designed to be fast, scalable, and highly performant, making it well-suited for large-scale machine learning applications. One of the key features of XGBoost is its use of regularisation functions to prevent overfitting and improve the generalisation of the model. XGBoost also uses a tree pruning algorithm to remove nodes with low feature importance to reduce the complexity of the model and improve accuracy. 

XGBoost has been highly successful for tabular data analysis, and deep learning researchers have been striving to surpass its performance.

\subsection{Attention and Transformers}

Transformers, originating from Vaswani et al. \cite{vaswani2017attention}, are a type of deep learning architecture originally developed for natural language processing tasks and have been adapted for use in the tabular domain. These models use self-attention to compute the importance of each feature within the context of the entire dataset, enabling them to learn complex, non-linear relationships between features. This is contrasted to GBDTs where all features are treated equally and relationships are not considered between them. Attention mechanisms are capable of highlighting relevant features and patterns in the dataset that are the most informative for making accurate predictions. 

Multi-head self-attention is a type of attention mechanism used in Transformers. A weight is assigned to each input token based on its relevance to the output, allowing selective focus on different parts of the input data. 

\begin{figure}[!t]
    \centering
    \includegraphics[]{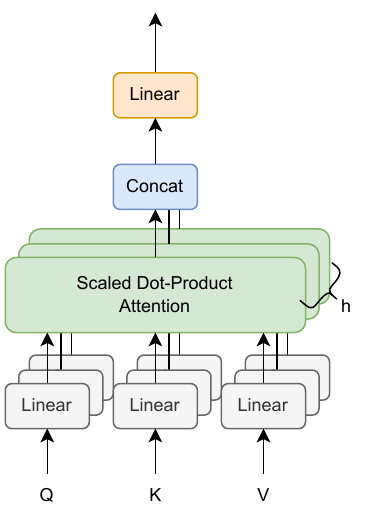}
    \caption{Multi-head attention from \cite{vaswani2017attention}, where h is the number of heads, Q, K, and V are the query, key and value vectors.}
    \label{fig:attention}
\end{figure}

The attention mechanism is applied multiple times in parallel, with each attention head attending to a different subspace of the input representation, allowing the model to capture different aspects of the input data and learn more complex, non-linear relationships between the inputs. The outputs of the multiple attention heads are then concatenated and passed through a linear layer to produce the final output. This is depicted in Figure \ref{fig:attention}, where the scaled dot-product attention is:

\begin{equation}
Attention(Q,K,V) = softmax(\frac{QK^T}{\sqrt{d_k}})V 
\label{eq:attention}
\end{equation}

In Figure \ref{fig:attention} and Equation \ref{eq:attention}, Q, K, and V are the query, key and value vectors used to compute attention weights between each element of the input sequence. $d_k$ is the dimension of the key vectors. 
 
SAINT \cite{somepalli2022saint}, the Self-Attention and Intersample Attention Transformer, is a deep learning model designed to make predictions based on tabular data. SAINT utilises attention to highlight specific features or patterns within the dataset that are most relevant for making accurate predictions, helping models better understand complex relationships within the data and make more accurate predictions. 

In their experiments, they find that SAINT, on average, outperforms all other methods on supervised and semi-supervised tasks for regression, including GBDT-based methods, on a variety of datasets.

\subsubsection{Chemical Kinetics}

Siemens Energy developed a Chemical Kinetics PEMS model \cite{hackney2016predictive} through mapping emissions via a 1D reactor element code 'GENE-AC' computational fluid dynamics model of their SGT-400 combustor and converting this to a parametric PEMS model. This is a first principles method that uses factors such as pilot/main fuel split, inlet air temperature and inlet air pressure to calculate the predicted emissions.

\section{Related Works}

\subsection{Gas Turbine Emissions Prediction}

\subsubsection{First Principles}

Predictive emissions monitoring systems (PEMS) for gas turbines have been developed since 1973 \cite{hung1975experimentally} in which an analytical model was developed using thermodynamics to predict NOx emissions. Rudolf et al. \cite{rudolf2016modelling} developed a mathematical model which takes into account performance deterioration due to engine ageing. They combined different datasets, such as validation measurements and long-term operational data, to provide more meaningful emission trends. Lipperheide et al. \cite{lipperheide2018long} also incorporate aging of the gas turbines into their analytical model which is capable of accurately predicting NOx emissions for power in the range 60-100\%. Siemens Energy developed a Chemical Kinetics model \cite{hackney2016predictive} to accurately predict CO and NOx emissions for their SGT-400 gas turbine. They used a 1D reactor model to find the sensitivity of the emissions to the different input parameters as a basis for the PEMS algorithm. Bainier et al. \cite{bainier2016two} monitor their analytical PEMS over two years and find a continuous good level of accuracy, noting that training is required to fully upkeep the system. 

\subsubsection{Machine Learning}

A number of machine learning (ML) methods have been used to predict emissions for gas turbines and have been found to be more flexible for prediction than first principles methods. Cuccu et al. \cite{cuccu2017data} compared twelve machine learning methods including linear regression, kernel based methods and feed-forward artificial neural networks with different backpropagation methods. They used k-fold cross-validation to select the optimal method-specific parameters and found that improved resilient backpropagation (iRPROP) achieved the best performance, and note that thorough pre-processing is required to produce such results. Kaya et al. \cite{kaya2019predicting} compared three decision fusion schemes on a novel gas turbine dataset, highlighting the importance of certain features within the dataset for prediction. Si et al. \cite{si2019development} also used k-fold validation to determine the optimal hyperparameters for their neural-network based models. Rezazadeh et al. \cite{rezazadeh2020environmental} proposed a k-nearest-neighbour algorithm to predict NOx emissions. 

Azzam et al. \cite{azzam2018application} utilised evolutionary artificial neural networks and support vector machines to model NOx emissions from gas turbines, finding that use of their genetic algorithm results in a high enough accuracy to offset the computational cost compared to the cheaper support vector machines. Kochueva et al. \cite{kochueva2021data} develop a model based on symbolic regression and a genetic algorithm with a fuzzy classification model to determine "standard" or "extreme" emissions levels to further improve their prediction model. Botros et al. \cite{botros2009verification, botros2010neural, botros2011predictive} developed a predictive emissions model based on neural networks with an accuracy of $\pm$10 parts per million.

Guo et al. \cite{guo2022nox} developed a NOx prediction model based on attention mechanisms, LSTM, and LightGBM. The attention mechanisms were introduced into the LSTM model to deal with the sequence length limitation LSTM faces. They eliminate noise through singular spectrum analysis and then use LightGBM to select the dependent feature. This processed data is then used as input to the LSTM and the attention mechanism is used to enhance the historical learning ability of information. They add feature attention and temporal attention to the LSTM model to improve prediction by allowing different emphasis by allocating different weights.

\subsection{Tabular Prediction}

\subsubsection{Tree-Based}

Gradient-boosted decision trees (GBDTs) have emerged as the dominant approach for tabular prediction, with deep learning methods only beginning to outperform them in some cases. Notably, XGBoost \cite{chen2016xgboost} often achieves state-of-the-art performance in regression problems. Other GBDTs such as LightGBM \cite{ke2017lightgbm} and CatBoost \cite{prokhorenkova2018catboost} have shown success in tabular prediction.

Deep learning faces challenges when dealing with tabular data, such as low-quality training data, the lack of spatial correlation between variables, dependency on preprocessing, and the impact of single features \cite{borisov2022deep}. Shwartz et al. \cite{shwartz2022tabular} conclude that deep models were weaker than XGBoost, and that deep models only outperformed XGBoost alone when used as an ensemble with XGBoost. They also highlight the challenges in optimizing deep models compared to XGBoost. Grinsztajn et al. \cite{grinsztajn2022tree} find that tree-based models are state-of-the-art on medium sized data (10,000 samples), especially when taking into account computational cost, due to the specific features of tabular data, such as uninformative features, non rotationally-invariant data, and irregular patterns in the target function. Kadra et al. \cite{kadra2021well} argue that well-regularized plain MLPs significantly outperform more specialized neural network architectures, even outperforming XGBoost. 

\subsubsection{Attention and Transformers}

Attention- and transformer-based methods have shown promise in recent years for tabular prediction. Ye et al. \cite{ye2022applying} provide an overview on attention-based approaches for tabular data, highlighting the benefits of attention in tabular models. SAINT \cite{somepalli2022saint} introduced intersample attention, which allows rows attend to each other, as well as using the standard self-attention mechanism, leading to improved performance over GBDTs on a number of benchmark tasks. TabNet \cite{arik2021tabnet} is an interpretable model that uses sequential attention to select features to reason from at each step. FT-Transformer \cite{gorishniy2021revisiting} is a simple adaption of the Transformer architecture that has outperformed other deep learning solutions on most tasks. However, GBDTs still outperform it on some tasks. TabTransformer \cite{huang2020tabtransformer} transforms categorical features into robust contextual embeddings using transformer layers, but it does not affect continuous variables. Kossen et al. \cite{kossen2021self} took the entire dataset as input and used self-attention to reason about relationships between datapoints. ExcelFormer \cite{chen2023excelformer} alternated between two attention modules to manipulate feature interactions and feature representation updates and manages to convincingly outperform GBDTs.

Despite the promising results of these attention- and transformer-based methods, deep learning models have generally been weaker than GBDTs on datasets that were not originally used in their respective papers \cite{shwartz2022tabular}. Proper pre-processing, pre-training \cite{rubachev2022revisiting} and embedding \cite{gorishniy2022embeddings} can enable deep learning tabular models to perform significantly better, reducing the gap between deep learning and GBDT models.

\section{Methodology}

\subsection{Data}

The data is test bed data from the Siemens SGT400 gas turbines. This is tabular data consisting of a number of different gas turbines tested over a wide range of operating conditions. In total, there are 37,204 rows of data with 183 features including process parameters such as temperatures and pressures, and the target emission variables NOx and CO. All data is numerical values.

\subsection{Pre-Processing}
\label{sec:pre-processing}

From the test bed dataset, two comparison sub-datasets were used: "Full" and "Cropped". The Cropped dataset consisted of a significant amount of filters pre-applied to the data by Siemens Energy for the Chemical Kinetics model, while the Full dataset had no filters applied. Standard pre-processing was applied to both sets of data including removing rows with missing data, removing negatives from emissions data, and removing liquid fuel data. Features with a significant number of missing rows were also removed. For the Full dataset, any features with more than 18,100 missing values were removed. Similarly, for the Cropped dataset, features with more than 3,000 missing values were removed. These threshold values were chosen to be greater than the number of missing values than the maximum number of missing values found in the emission columns.

\begin{table}[t]
    \caption{Pre-processing process for the Full and Cropped datasets showing number of rows in each dataset.}
    \centering
    \label{tab:datasets}
    \begin{tabular}{{c|c|c}}
    \hline
    \textbf{Action} &  \textbf{Full}  & \textbf{Cropped} \\ 
    \hline
    Start & 37204 rows, 183 features & 9873 rows, 183 features\\ 
    Remove low data features  & Removes 9 features & Removes 95 features \\ 
    Remove liquid fuel data   & Removes 5752 rows     & No change    \\ 
    Remove negative emissions & Removes 16977 rows   & Removes 744 rows  \\ 
    Remove all missing values & Removes 8615 rows  & Removes 2700 rows   \\ 
    End     & 5860 rows, 174 features   & 6429 rows, 88 features    \\ 
    \hline
    \end{tabular}
\end{table}

\begin{figure}[b]
    \centering
    \includegraphics[trim=0mm 0mm 0mm 0mm, width=\textwidth]{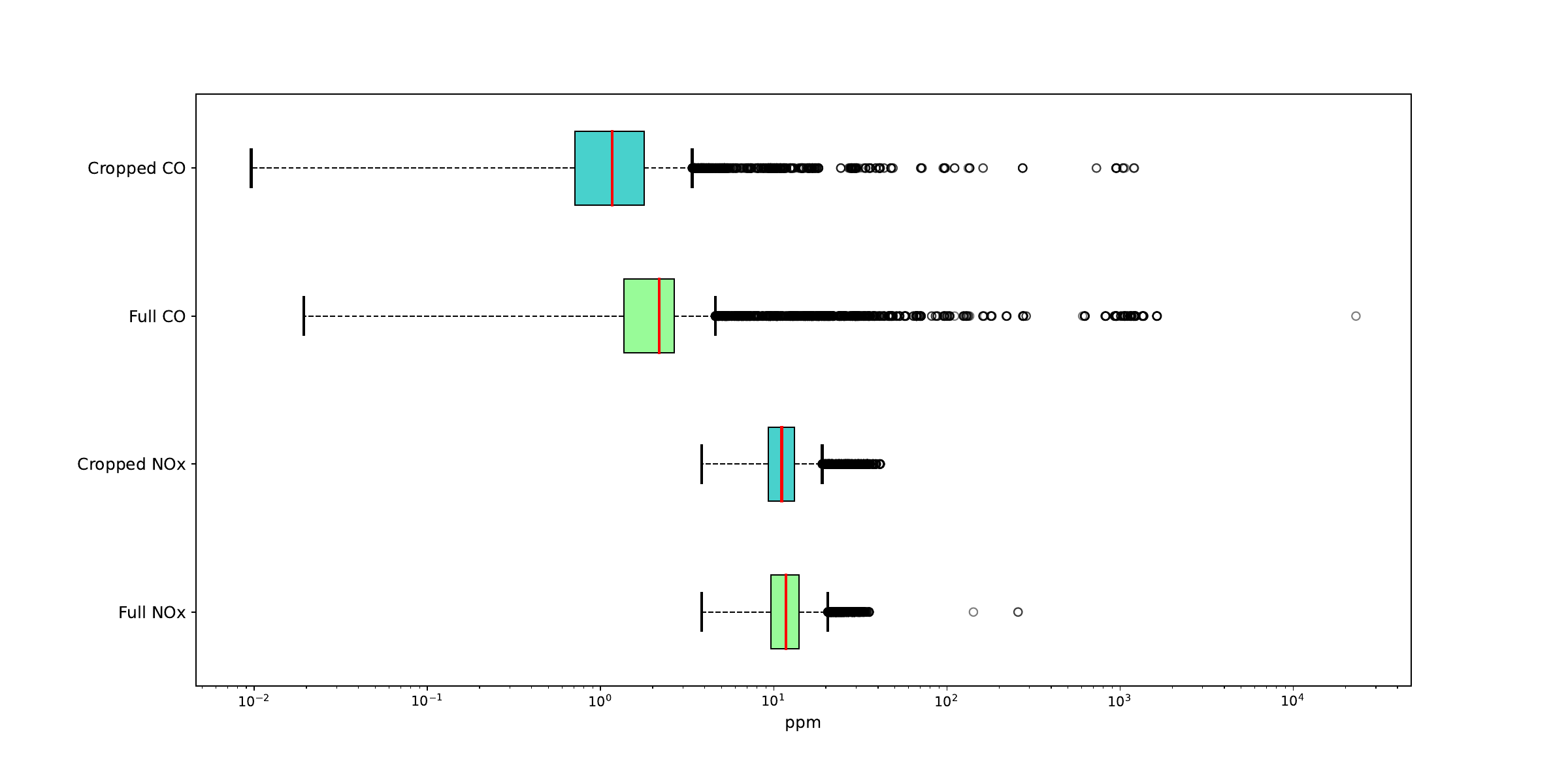}
    \caption{NOx and CO data spread for Full and Cropped datasets on a logarithmic scale.}
    \label{fig:nox_co_box}
\end{figure}

Table \ref{tab:datasets} provides an overview of both sub-datasets and the number of rows and features in each. Due to the prior pre-processing removing proportionally more missing values through the original filters, the Cropped dataset ends with more rows of data compared to the Full dataset, at the cost of reducing the number of features. When removing the same features from the Cropped dataset as the Full dataset only 2044 rows remain so this was not chosen to be used for modelling. Further feature details can be found in Table \ref{tab:all_features}.

The dataset is collected from 0\% to 126\% load, and pre-processing reduces this to 24\% to 126\%. We utilise this full range for our comparisons.

Figure \ref{fig:nox_co_box} depicts the spread of the data for the target emissions, NOx and CO, for both sub-datasets. CO has many more outliers compared to NOx, with some particularly far from the median.

\subsection{Models}

We compare a transformer-based model, SAINT \cite{somepalli2022saint}, and GBDT XGBoost \cite{chen2016xgboost}, against the an existing PEMS model used by Siemens Energy, a first principles-based Chemical Kinetics model \cite{hackney2016predictive}. 

\subsubsection{SAINT}

\begin{figure}[!t]
    \centering
    \includegraphics[width=\linewidth]{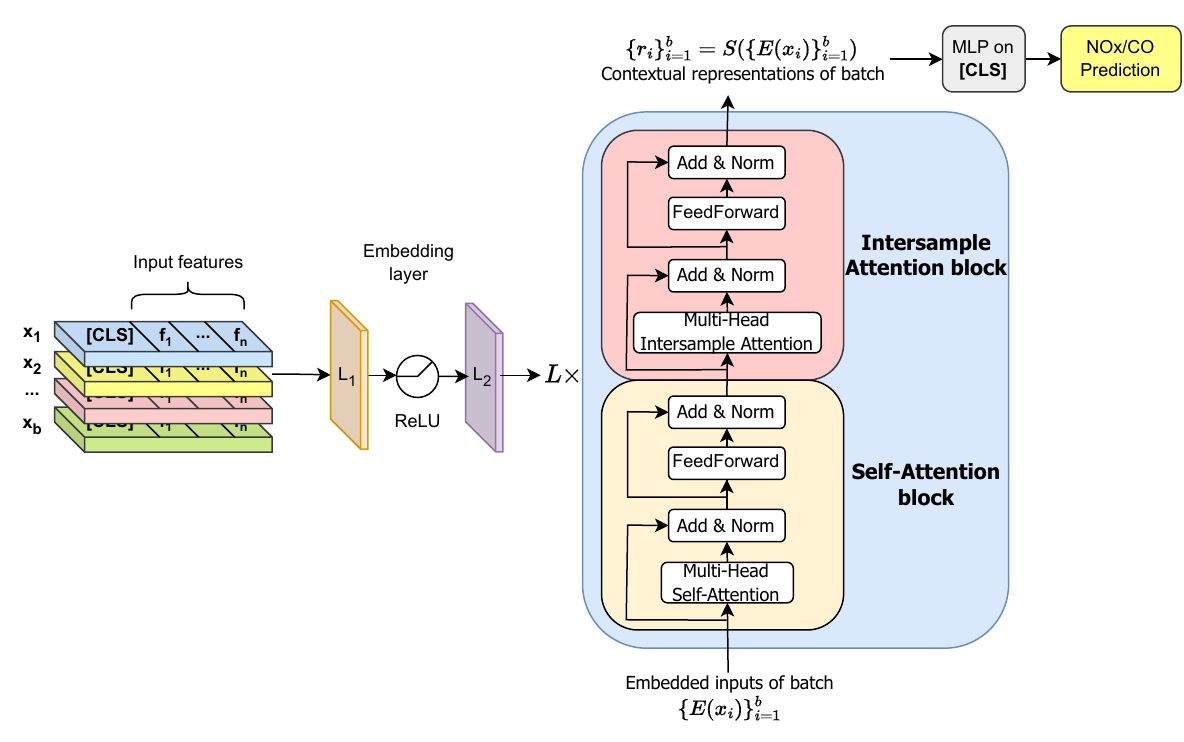}
    \caption{Proposed method based on SAINT \cite{somepalli2022saint}. The features, [$f_1, ..., f_n$], are the process parameters from sensors within the gas turbine tests, where $n$ is the number of features, 11. Each $x_i$ is one row of data including one of each feature, where $b$ is the batch size, 32. A [CLS] token with a learned embedding is appended to each data sample. This batch of inputs is passed through an embedding layer, consisting of a linear layer, a ReLU non-linearity, followed by a linear layer, prior to being processed by the SAINT model $L$ times, where $L$ is 3. Only representations corresponding to the [CLS] token are selected for an MLP to be applied to. MSE loss is done on predictions during training. For our experiments, $b$ is the batch size (32), $n$ is the number of features (7). $L_1$ is the first linear layer, with 1 input feature and 100 output features, $L_2$ is the second linear layer, with 100 input features and 1 output feature. The embedding layer is performed for each feature.}
    \label{fig:saint}
\end{figure}

SAINT accepts a sequence of feature embeddings as input and produces contextual representations with the same dimensionality. 

Features are projected into a combined dense vector space and passed as tokens into a transformer encoder. A single fully-connected layer with a ReLU activation is used for each continuous feature's embedding. 

SAINT alternates self-attention and intersample attention mechanisms to enable the model to attend to information over both rows and columns. The self-attention attends to individual features within each data sample, and intersample attention relates each row to other rows in the input, allowing all features from different samples to communicate with each other. 

Similar to the original transformer \cite{vaswani2017attention}, there are $L$ identical layers, each containing one self-attention and one intersample attention transformer block. The self-attention block is identical to the encoder from \cite{vaswani2017attention}, consisting of a multi-head self-attention layer with 8 heads, and two fully-connected feed-forward layers with a GELU non-linearity. A skip connection and layer normalisation are applied to each layer. The self-attention layer is replaced by an intersample attention layer for the intersample attention block. For the intersample attention layer, the embeddings of each feature are concatenated for each row, and attention is computed over samples rather than features, allowing communication between samples. 

We use SAINT in a fully supervised multivariate regression setting, which was not originally reported on in the paper. The code we based our experiments on can be found at\footnote{https://github.com/somepago/saint}. We used the AdamW optimiser with a learning rate of 0.0001.

\subsubsection{XGBoost}

XGBoost reduces overfitting through regularisation and pruning, using a distributed gradient boosting algorithm to optimise the model's objective function to make it more scalable and efficient, and automatically handles missing values. 

Decision trees are constructed in a greedy manner as a weak learner. At each iteration, XGBoost evaluates the performance of the current ensemble and adds a new tree to the ensemble that minimises the loss function through gradient descent. Each successive tree implemented compensates for residual errors in the previous tree. 

\begin{figure}[!t]
    \centering
    \includegraphics[trim=0mm 0mm 0mm 0mm, width=\textwidth]{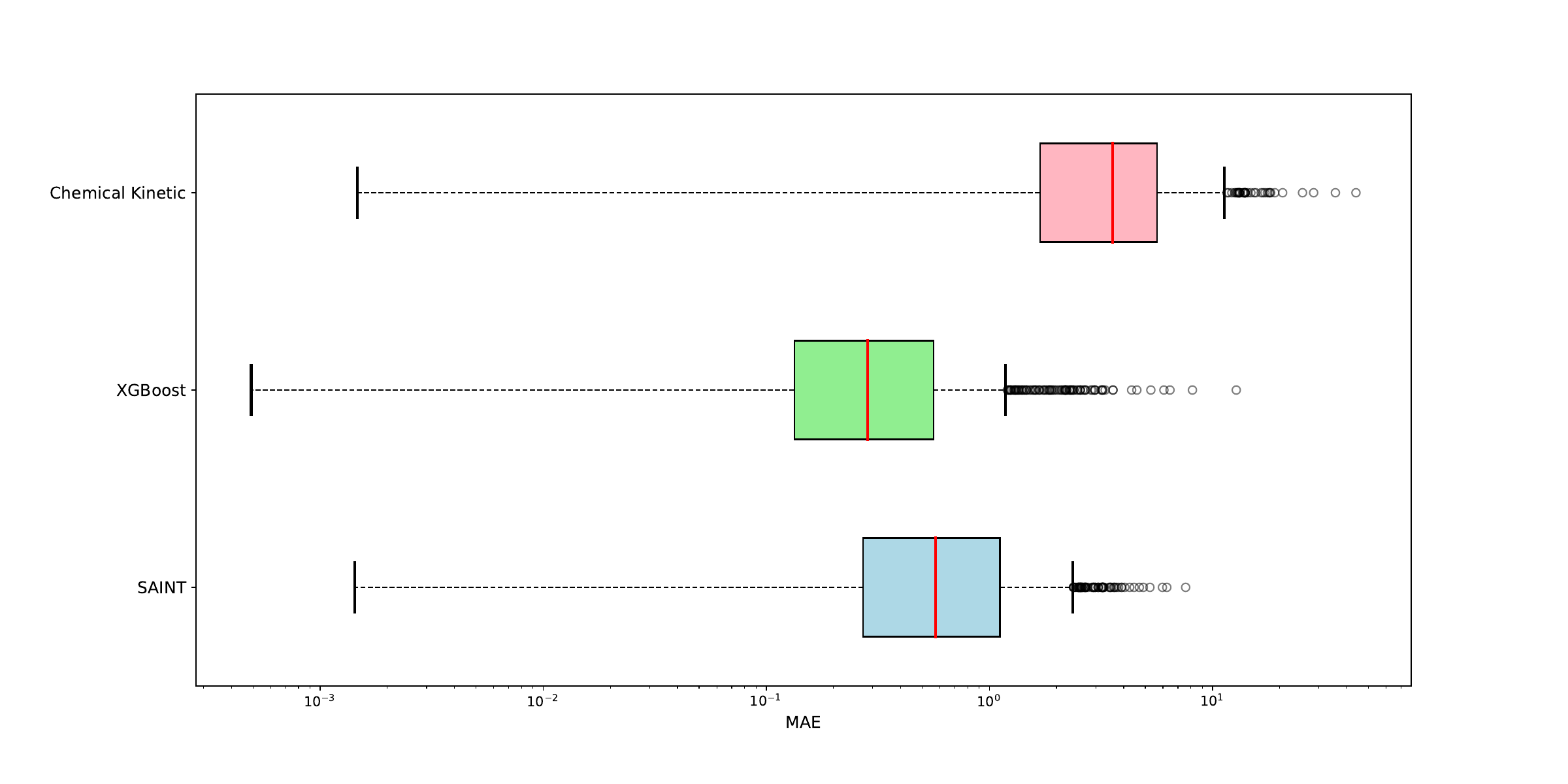}
    \caption{Box plots for MAE results for NOx for each model on a logarithmic scale.}
    \label{fig:nox_box}
\end{figure}

\subsubsection{Chemical Kinetics}

We compare our work to an updated Chemical Kinetics model, based on \cite{hackney2016predictive}, using the same sets of test data for comparisons. The predictions for the Chemical Kinetics model are essentially part of the original dataset, with the number of features and rows of each sub-dataset, described in Section \ref{sec:pre-processing}, not affecting the raw predictions but eliminating the varying rows depending on missing values due to features in the dataset.

\subsection{Metrics and Evaluation}

The metrics used to evaluate the models in this work are the mean absolute error (MAE) and root mean squared error (RMSE). 

MAE is expressed as follows: 

\begin{equation}
    MAE = \frac{1}{n} \sum \limits_{i=1}^n | y_i - \hat{y_i} | 
\end{equation}

RMSE is expressed as follows:

\begin{equation}
    RMSE = \sqrt{ \frac{1}{n} \sum\limits_{i=1}^n ( y_i - \hat{y_i} )^2 }
\end{equation}

We used randomised cross-validation to evaluate the performance of the machine learning models, SAINT and XGBoost, whereby the data was randomly sub-sampled 10 times to obtain unbiased estimates of the models' performance on new, unseen data on which they were re-trained and tested on. We report the average and standard deviation of the MAE and RMSE for each sub-dataset, providing an insight into the models' consistency and variation in performance. The Chemical Kinetics model is also compared on these test sets to provide relative benchmark for the performance of the models. The CO and NOx emissions targets are individually trained for to achieve specialised models for each target.

\subsection{Impact of Number of Features}

To assess the influence of the number of features compared to the number of rows of data on prediction performance, we further split each dataset where each subset contained a decreasing number of features, leading to fewer rows of missing data, allowing an examination into the effect of removing less important features on the availability of data points for training. Feature removal followed the order of decreasing feature importance according to XGBoost, where the importance is calculated by XGBoost based on how often each feature is used to make key decisions across all trees in the ensemble. The order of importance for each feature can be found in Table \ref{tab:all_features}.

\begin{figure}[!t]
    \centering
    \includegraphics[trim=0mm 0mm 0mm 0mm, width=\textwidth]{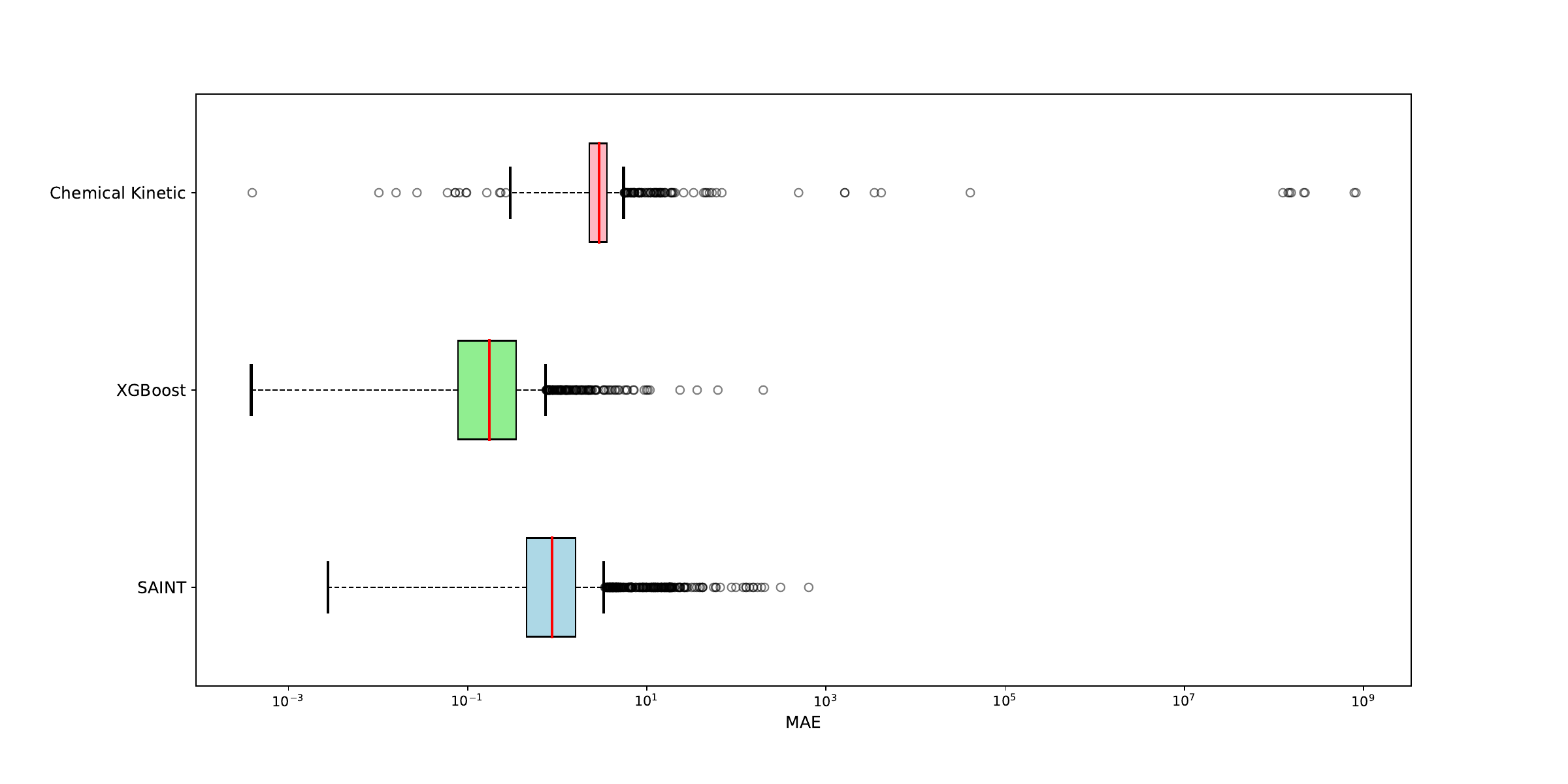}
    \caption{Box plots for MAE results for CO for each model on a logarithmic scale.}
    \label{fig:co_box}
\end{figure}

\section{Results and Discussion}

\begin{table*}[b]
\centering
\caption{Tabular prediction results for each model on the two sets of data and four sets of number of features used. Mean value for 10 dataset subsamples provided, with standard deviation in brackets.}
\label{tab:all_results}
\fontsize{7pt}{7pt}\selectfont
\begin{tabular}{c|c|cc|cc|cc}
\toprule[1.0pt]
\multicolumn{2}{c}{Methods}     & \multicolumn{2}{|c}{SAINT} & \multicolumn{2}{|c}{XGBoost} & \multicolumn{2}{|c}{Chemical Kinetic}  \\
\midrule[0.5pt]
\multicolumn{2}{c|}{Metric} & MAE & RMSE & MAE & RMSE & MAE & RMSE  \\

\midrule[1.0pt]

\multirow{4}{*}{\rotatebox{90}{\makecell{NOx \\ Full}}} & 174  & 0.91 $\pm$0.11 & \textbf{2.82 $\pm$2.45} & \textbf{0.62 $\pm$0.14} & 4.08 $\pm$3.09 &  4.46 $\pm$0.15 & 6.59 $\pm$1.43 \\

& 130  & 0.89 $\pm$0.21 & \textbf{2.92 $\pm$2.02} & \textbf{0.74 $\pm$0.18} & 4.48 $\pm$3.65 & 4.09 $\pm$0.10 & 6.14 $\pm$1.14 \\

& 87 & 1.72 $\pm$0.70 & \textbf{3.83 $\pm$1.62} & \textbf{0.76 $\pm$0.12} & 4.04 $\pm$2.62 & 4.09 $\pm$0.10 & 6.14 $\pm$1.14 \\

& 45 & 1.14 $\pm$0.38 & \textbf{2.96 $\pm$1.64} & \textbf{0.74 $\pm$0.08} & 3.00 $\pm$1.99 & 3.68 $\pm$0.12 & 5.55 $\pm$0.94 \\

\midrule[0.5pt]

\multirow{4}{*}{\rotatebox{90}{\makecell{NOx \\ Cropped}}} & & & & & & & \\  

& 88 & 0.54 $\pm$0.08 & \textbf{0.92 $\pm$0.1} & \textbf{0.47 $\pm$0.02} & 0.95 $\pm$0.17 & 2.67 $\pm$0.06 & 3.84 $\pm$0.33\\

& 45  & 0.56 $\pm$0.07 & 0.94 $\pm$0.07 &\textbf{0.44 $\pm$0.02} & \textbf{0.92 $\pm$0.16} & 2.67 $\pm$0.06 & 3.84 $\pm$0.33 \\

& & & & & & & \\

\midrule[2.0pt]

\multirow{4}{*}{\rotatebox{90}{\makecell{CO \\ Full}}} & 174  & 11.37 $\pm$6.61 & \textbf{117.61 $\pm$191.07} & \textbf{5.05 $\pm$6.45} & 117.83 $\pm$197.50 & 2.49E+6 $\pm$7.54E+5 & 3.79E+7 $\pm$7.35E+6 \\    

& 130 & 10.58 $\pm$5.84 & \textbf{164.20 $\pm$225.07} & \textbf{7.41 $\pm$8.09} & 220.53 $\pm$260.67 & 1.47E+6 $\pm$5.98E+5 & 2.85E+7 $\pm$7.37E+6 \\  

& 87 & 14.31 $\pm$6.33 & \textbf{152.70 $\pm$225.24} & \textbf{7.68 $\pm$10.80} & 214.44 $\pm$317.08 & 1.50E+6 $\pm$5.98E+5 & 2.85E+7 $\pm$7.37E+6 \\

& 45 & 24.97 $\pm$30.58 & 292.55 $\pm$236.71 & \textbf{6.04 $\pm$6.30} & \textbf{219.92 $\pm$262.52} & 1.38E+6 $\pm$8.93E+5 & 2.64E+7 $\pm$1.28E+7 \\

\midrule[0.5pt]
\multirow{4}{*}{\rotatebox{90}{\makecell{CO \\ Cropped}}} & & & & & & & \\  

& 88 & 2.46 $\pm$0.72 & 20.02 $\pm$10.14 & \textbf{0.59 $\pm$0.31} & \textbf{9.13 $\pm$8.15} & 5.97E+5 $\pm$3.32E+5 & 1.80E+7 $\pm$9.34E+6 \\ 

& 45  & 2.73 $\pm$2.30 & 20.01 $\pm$10.15 & \textbf{0.63 $\pm$0.37} & \textbf{10.50 $\pm$9.31} & 5.96E+5 $\pm$3.32E+5 & 1.80E+7 $\pm$9.34E+6 \\

& & & & & & & \\  

\bottomrule[1.0pt]

\end{tabular}
\end{table*}

Table \ref{tab:all_results} describes the average MAE and RMSE obtained from the 10 sub-samples of the dataset with the varying number of features. XGBoost has on average the lowest MAE for each emission and number of features, while SAINT has a lower RMSE on average. 

All models, especially the Chemical Kinetics model, have significant errors when predicting CO. Further analysis of these results indicated that these large errors were primarily driven by a small number of data points with extremely anomalous MAE values. Figure \ref{fig:co_box} illustrates these outliers, with the logarithmic scale emphasizing the limited number of data points responsible for the higher mean MAE. Despite the presence of outliers, the median MAE values for each model were not excessively high, with the majority of data points exhibiting more accurate predictions for CO.

\begin{figure}[!t]
    \centering
    \includegraphics[trim=0mm 0mm 0mm 0mm, width=\textwidth]{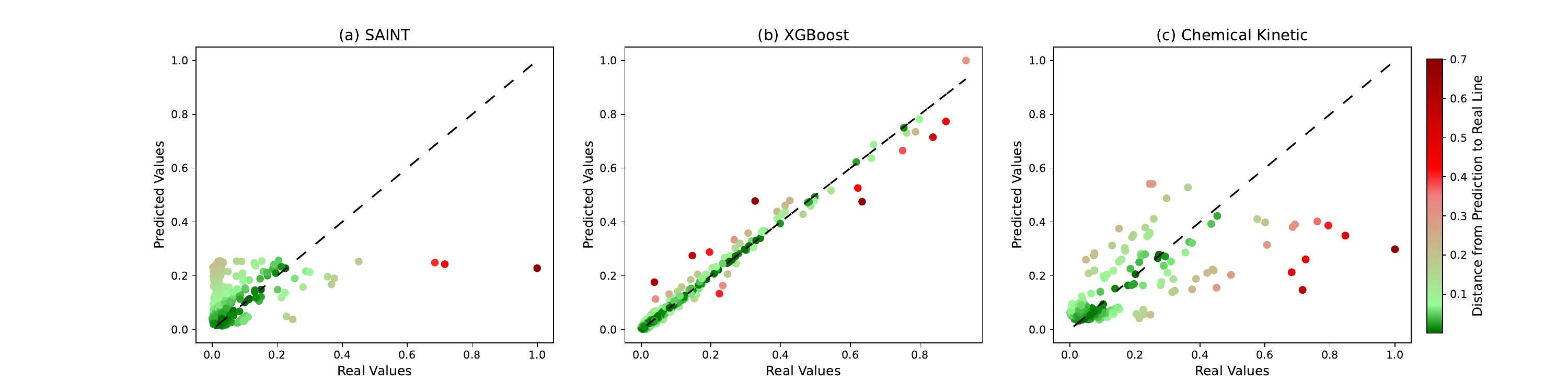}
    \caption{Normalised real vs. predicted values for CO for each model within one standard deviation for the Full dataset with all features. Extreme anomalous real and predicted values above 1000 were also removed, removing 14 data points.}
    \label{fig:CO_preds_vs_real}
\end{figure}

Figure \ref{fig:CO_preds_vs_real} demonstrates that the majority of predictions generated by all models fall within a reasonable range for accurate CO emission prediction for gas turbines. While overall performance may be affected by the presence of outliers, the models do exhibit good predictive capabilities for CO and NOx emissions.

\begin{figure}[!t]
    \centering
    \includegraphics[trim=0mm 0mm 0mm 0mm, width=\textwidth]{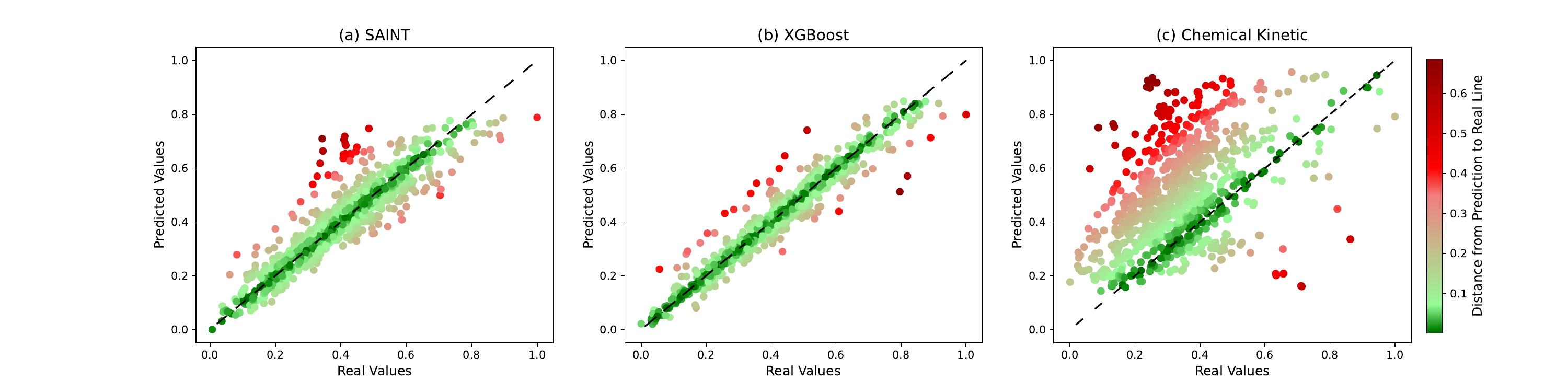}
    \caption{Normalised real vs. predicted values for NOx for each model within one standard deviation.}
    \label{fig:NOx_preds_vs_real}
\end{figure}

Figure \ref{fig:NOx_preds_vs_real} and \ref{fig:CO_preds_vs_real} show the normalised predictions compared to the real value for NOx and CO. For Figure \ref{fig:CO_preds_vs_real}, the predictions above 1000ppm were removed from view as these were extremely anomalous and prevented the main results to be seen clearly. For both emissions, the Chemical Kinetics model has more spread compared to SAINT and XGBoost. For CO especially, XGBoost predictions are closer to the identity line compared to SAINT.

\begin{figure}[!t]
    \centering
    \includegraphics[trim=0mm 0mm 0mm 0mm, width=\textwidth]{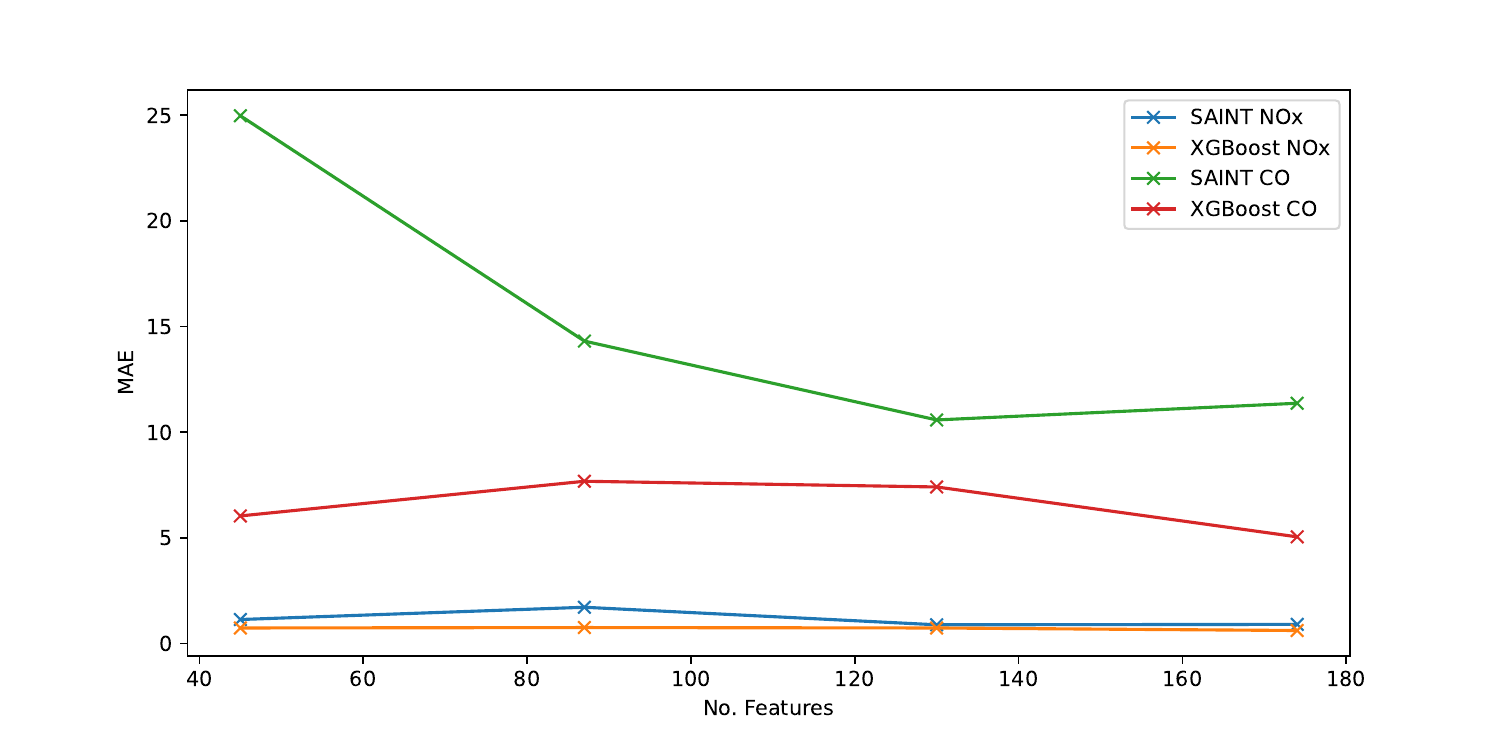}
    \caption{MAE compared to number of features for the Full dataset. For training, on average between the 10 sub-datasets, 174 features had 3808 rows, 130 and 87 features had 5084 rows, 45 features had 6223 rows.}
    \label{fig:mae_vs_feats}
\end{figure}

Figure \ref{fig:mae_vs_feats} displays the relationship between the MAE values and the number of features in the analysis, highlighting the potential impact of feature removal and its effect on prediction performance. Despite having 2415 more rows of training data, with the exception of SAINT's CO prediction, the MAE is not significantly affected by the change in number of features and rows.

In our evaluation, XGBoost provided the best prediction accuracy for both NOx and CO, with both machine learning methods outperforming the original Chemical Kinetics model. Prediction for NOx is significantly more accurate than CO prediction for all models. This can be attributed to the wider spread of data points and greater presence of influential outliers in the CO real values, as evident in Figure \ref{fig:nox_co_box}. The abundance of outliers in the CO dataset made it inherently more challenging to predict accurately. The filters used for the Cropped dataset particularly improved the RMSE of the machine learning models as it removed some outlier inputs in the dataset.

\section{Conclusion and Future Work}

XGBoost remains the best model for tabular prediction for this gas turbine dataset for both NOx and CO, but the attention-based model, SAINT, is catching up in terms of performance. Both machine learning models outperformed the first-principles-based Chemical Kinetics model, indicating that machine learning continues to show a promising future for gas turbine emissions prediction.

Furthermore, to fully utilise the years of operational gas turbine data that is available but unlabelled, a future step to improve gas turbine emissions prediction will be to include self-supervised learning into the training process. Despite XGBoost displaying the best performance here, attention-based methods such as SAINT will be easier to combine with self-supervised learning by performing a pretext task such as masking to predict masked sections of the operational data to learn representations of the data, which can then be used in a downstream task using SAINT to create predictions.

\section*{Aknowledgements}
The work presented here received funding from EPSRC (EP/W522089/1) and Siemens Energy Industrial Turbomachinery Ltd. as part of the iCASE EPSRC PhD studentship "Predictive Emission Monitoring Systems for Gas Turbines".

 \bibliographystyle{elsarticle-num} 
 \bibliography{references}

\appendix

\section{Appendix}
\setcounter{table}{0}

\begin{longtable}{|p{6cm}|p{2cm}|p{2cm}|p{2cm}|p{2.2cm}|}
\captionsetup{width=.96\textwidth}
\caption{Complete list of features in test bed dataset ordered from least to most missing values with XGBoost importance order for Full and Cropped sub-datasets. Lower values indicate highest XGBoost importance for the final model proposed.}
\label{tab:all_features}
\centering
\endfirsthead
\hline
Description & Unit & Missing Values & Full Importance & Cropped Importance 
\\ \hline
\endhead
\hline
Description & Unit & Missing Values & Full Importance & Cropped Importance 
\\ \hline
Compressor exit pressure & barg & 6 & 0 & 80 \\ \hline
Turbine interduct temperature    & \degree C    & 6   & 1    & 2 \\ \hline
Pressure drop across exhaust ducting  & mbar   & 6   & 4    & 70\\ \hline
Exhaust temperature  & \degree C    & 6   & 5    & 51\\ \hline
Turbine interduct temperature    & \degree C    & 6   & 6    & 5 \\ \hline
Turbine interduct temperature    & \degree C    & 6   & 7    & 23\\ \hline
Power turbine shaft speed  & rpm    & 6   & 18   & 76\\ \hline
Turbine interduct temperature    & \degree C    & 6   & 20   & 7 \\ \hline
Pressure drop across inlet ducting     & mbar   & 6   & 21   & 11\\ \hline
Exhaust temperature  & \degree C    & 6   & 24   & 64\\ \hline
Turbine interduct temperature    & \degree C    & 6   & 37   & 34\\ \hline
Temperature after inlet ducting  & \degree C    & 6   & 38   & 21\\ \hline
Temperature after inlet ducting  & \degree C    & 6   & 39   & 62\\ \hline
Turbine interduct temperature    & \degree C    & 6   & 49   & 44\\ \hline
Exhaust temperature  & \degree C    & 6   & 58   & 67\\ \hline
Exhaust temperature  & \degree C    & 6   & 59   & 24\\ \hline
Exhaust temperature  & \degree C    & 6   & 74   & 27\\ \hline
Compressor shaft speed     & rpm    & 6   & 78   & 12\\ \hline
Turbine interduct temperature    & \degree C    & 6   & 82   & 19\\ \hline
Exhaust temperature  & \degree C    & 6   & 83   & 75\\ \hline
Exhaust temperature  & \degree C    & 6   & 90   & 18\\ \hline
Exhaust temperature  & \degree C    & 6   & 91   & 41\\ \hline
Exhaust temperature  & \degree C    & 6   & 96   & 74\\ \hline
Temperature in filter house (ambient temperature) & \degree C    & 6   & 110  & 54\\ \hline
Exhaust temperature  & \degree C    & 6   & 111  & 86\\ \hline
Compressor exit temperature& \degree C    & 6   & 112  & 68\\ \hline
Turbine interduct temperature    & \degree C    & 6   & 114  & 52\\ \hline
Compressor exit temperature& \degree C    & 6   & 115  & 13\\ \hline
Exhaust temperature  & \degree C    & 6   & 125  & 48\\ \hline
Turbine interduct temperature    & \degree C    & 6   & 126  & 56\\ \hline
Temperature after inlet ducting  & \degree C    & 6   & 147  & 38\\ \hline
Turbine interduct temperature    & \degree C    & 6   & 149  & 16\\ \hline
Exhaust temperature  & \degree C    & 6   & 150  & 79\\ \hline
Turbine interduct temperature    & \degree C    & 6   & 153  & 25\\ \hline
Turbine interduct pressure & barg    & 6   & 156  & 15\\ \hline
Turbine interduct temperature    & \degree C    & 6   & 159  & 26\\ \hline
Exhaust temperature  & \degree C    & 6   & 163  & 87\\ \hline
Turbine interduct temperature    & \degree C    & 6   & 171  & 58\\ \hline
Temperature after inlet ducting  & \degree C    & 23  & 32   & 30\\ \hline
Ambient pressure     & bara & 33  & 40   & 49\\ \hline
Temperature after inlet ducting  & \degree C    & 50  & 105  & 22\\ \hline
Variable guide vanes position    &  & 58  & 3    & 39\\ \hline
Temperature after inlet ducting  & \degree C    & 88  & 36   & 28\\ \hline
Inlet air mass flow  & kg/s   & 214 & 41   & 43\\ \hline
Turbine inlet pressure     & Pa     & 219 & 22   & 82\\ \hline
Fuel mass flow & kg/s   & 219 & 27   & 84\\ \hline
Calculated heat input (fuel flow method)    & W& 219 & 33   & 72\\ \hline
Turbine inlet temperature  & K& 219 & 35   & 6 \\ \hline
Mass flow into combustor (after bleeds)     & kg/s   & 219 & 66   &   \\ \hline
Power    & MW     & 219 & 109  & 83\\ \hline
Calculated heat input (heat balance method) & W& 219 & 123  & 47\\ \hline
Exhaust mass flow    & kg/s   & 219 & 151  & 66\\ \hline
Bleed mass flow& kg/s   & 219 & 68   & 65\\ \hline
Lower calorific value of fuel    & kJ/kg  & 468 & 162  & 37\\ \hline
Combustor 2 pilot-tip temperature& \degree C    & 970 & 12   & 1 \\ \hline
Combustor 4 pilot-tip temperature& \degree C    & 970 & 14   & 3 \\ \hline
Combustor 6 pilot-tip temperature& \degree C    & 970 & 29   & 8 \\ \hline
Combustor 5 pilot-tip temperature& \degree C    & 970 & 106  & 4 \\ \hline
Combustor 1 pilot-tip temperature& \degree C    & 970 & 121  & 36\\ \hline
Combustor 3 pilot-tip temperature& \degree C    & 970 & 127  & 14\\ \hline
Firing temperature   & K& 2178& 79   & 42\\ \hline
Load \% 1& \%     & 2837& 46   & 78\\ \hline
Load \% 2& \%     & 2837& 30   & 59\\ \hline
Bleed valve angle    & \%     & 2837& 26   & 85\\ \hline
Main/pilot burner split    & \%     & 3806& 102  & 10\\ \hline
Fuel demand    & kW     & 3806& 119  & 40\\ \hline
Main/pilot burner split    & \%     & 3806& 168  & 0 \\ \hline
Bleed valve angle    & Degrees& 3854& 154  & 9 \\ \hline
Gas Generator inlet journal bearing temperature 2 & \degree C    & 4172& 10   & 46\\ \hline
Gas Generator exit journal bearing temperature 2  & \degree C    & 4172& 70   & 57\\ \hline
Gas Generator Thrust Bearing temperature 2  & \degree C    & 4172& 73   & 20\\ \hline
Gas Generator Thrust Bearing temperature 1  & \degree C    & 4172& 113  & 63\\ \hline
Power Turbine Thrust Bearing temperature 2  & \degree C    & 4597& 64   & 29\\ \hline
Power Turbine exit journal bearing temperature 2  & \degree C    & 4597& 80   & 31\\ \hline
Power Turbine Thrust Bearing temperature 1  & \degree C    & 4597& 88   & 35\\ \hline
Power Turbine inlet journal bearing temperature 1 & \degree C    & 4597& 140  & 32\\ \hline
Compressor exit pressure   & bara & 8973&&   \\ \hline
Gas Generator inlet journal bearing temperature 1 & \degree C    & 9389& 77   & 45\\ \hline
Gas Generator exit journal bearing temperature 1  & \degree C    & 9389& 144  & 71\\ \hline
Power Turbine Exit Journal Y     & \textmu m     & 9814& 8    & 55\\ \hline
Power Turbine Exit Journal X     & \textmu m     & 9814& 11   & 50\\ \hline
Gas Generator Exit Journal Y     & \textmu m     & 9814& 13   & 81\\ \hline
Power Turbine Inlet Journal Y    & \textmu m     & 9814& 28   & 69\\ \hline
Power Turbine exit journal bearing temperature 1  & \degree C    & 9814& 69   & 33\\ \hline
Gas Generator Exit Journal X     & \textmu m     & 9814& 75   & 73\\ \hline
Power Turbine Inlet Journal X& \textmu m     & 9814& 87   & 77\\ \hline
Gas Generator Inlet Journal X    & \textmu m     & 9814& 101  & 53\\ \hline
Gas Generator Inlet Journal Y    & \textmu m     & 9814& 120  & 60\\ \hline
Power Turbine inlet journal bearing temperature 2 & \degree C    & 9814& 141  & 61\\ \hline
Combustor can 3, magnitude in second peak frequency in band 2 & psi    & 15020     & 2    &   \\ \hline
Combustor can 1, second peak frequency in band 1  & hz     & 15020     & 9    &   \\ \hline
Combustor can 3, magnitude in third peak frequency in band 2  & psi    & 15020     & 15   &   \\ \hline
Combustor can 5, magnitude in first peak frequency in band 2  & psi    & 15020     & 16   &   \\ \hline
Combustor can 1, first peak frequency in band 1   & hz     & 15020     & 17   &   \\ \hline
Combustor can 6, magnitude in first peak frequency in band 1  & psi    & 15020     & 23   &   \\ \hline
Combustor can 2, first peak frequency in band 2   & hz     & 15020     & 25   &   \\ \hline
Combustor can 2, first peak frequency in band 1   & hz     & 15020     & 31   &   \\ \hline
Combustor can 5, first peak frequency in band 1   & hz     & 15020     & 42   &   \\ \hline
Combustor can 4, magnitude in first peak frequency in band 2  & psi    & 15020     & 43   &   \\ \hline
Combustor can 4, third peak frequency in band 2   & hz     & 15020     & 44   &   \\ \hline
Combustor can 1, magnitude inthird peak frequency in band 2   & psi    & 15020     & 45   &   \\ \hline
Combustor can 3, first peak frequency in band 2   & hz     & 15020     & 47   &   \\ \hline
Combustor can 4, magnitude in third peak frequency in band 2  & psi    & 15020     & 50   &   \\ \hline
Combustor can 1, third peak frequency in band 2   & hz     & 15020     & 54   &   \\ \hline
Combustor can 6, magnitude in second peak frequency in band 2 & psi    & 15020     & 55   &   \\ \hline
Combustor can 6, first peak frequency in band 2   & hz     & 15020     & 62   &   \\ \hline
Combustor can 3, magnitude in first peak frequency in band 2  & psi    & 15020     & 63   &   \\ \hline
Combustor can 4, second peak frequency in band 2  & hz     & 15020     & 65   &   \\ \hline
Combustor can 2, second peak frequency in band 1  & hz     & 15020     & 67   &   \\ \hline
Combustor can 1, second peak frequency in band 2  & hz     & 15020     & 71   &   \\ \hline
Combustor can 5, magnitude in third peak frequency in band 2  & psi    & 15020     & 72   &   \\ \hline
Combustor can 2, third peak frequency in band 2   & hz     & 15020     & 76   &   \\ \hline
Combustor can 5, magnitude in first peak frequency in band 1  & psi    & 15020     & 81   &   \\ \hline
Combustor can 6, second peak frequency in band 2  & hz     & 15020     & 89   &   \\ \hline
Combustor can 4, magnitude in second peak frequency in band 2 & psi    & 15020     & 94   &   \\ \hline
Combustor can 2, magnitude in first peak frequency in band 1  & psi    & 15020     & 95   &   \\ \hline
Combustor can 5, third peak frequency in band 2   & hz     & 15020     & 97   &   \\ \hline
Combustor can 1, magnitude in second peak frequency in band 1 & psi    & 15020     & 98   &   \\ \hline
Combustor can 3, magnitude in first peak frequency in band 1  & psi    & 15020     & 99   &   \\ \hline
Combustor can 6, first peak frequency in band 1   & hz     & 15020     & 100  &   \\ \hline
Combustor can 3, second peak frequency in band 1  & hz     & 15020     & 104  &   \\ \hline
Combustor can 3, magnitude in second peak frequency in band 1 & psi    & 15020     & 107  &   \\ \hline
Combustor can 2, magnitude in second peak frequency in band 2 & psi    & 15020     & 108  &   \\ \hline
Combustor can 5, second peak frequency in band 2  & hz     & 15020     & 116  &   \\ \hline
Combustor can 4, magnitude in second peak frequency in band 1 & psi    & 15020     & 117  &   \\ \hline
Combustor can 5, first peak frequency in band 2   & hz     & 15020     & 118  &   \\ \hline
Combustor can 4, magnitude in first peak frequency in band 1  & psi    & 15020     & 129  &   \\ \hline
Combustor can 1, magnitude in first peak frequency in band 2  & psi    & 15020     & 130  &   \\ \hline
Combustor can 6, magnitude in first peak frequency in band 2  & psi    & 15020     & 132  &   \\ \hline
Combustor can 6, magnitude in third peak frequency in band 2  & psi    & 15020     & 133  &   \\ \hline
Combustor can 1, first peak frequency in band 2   & hz     & 15020     & 134  &   \\ \hline
Combustor can 2, magnitude in third peak frequency in band 2  & psi    & 15020     & 135  &   \\ \hline
Combustor can 6, third peak frequency in band 2   & hz     & 15020     & 136  &   \\ \hline
Combustor can 5, magnitude in second peak frequency in band 2 & psi    & 15020     & 143  &   \\ \hline
Combustor can 3, second peak frequency in band 2  & hz     & 15020     & 145  &   \\ \hline
Combustor can 4, first peak frequency in band 2   & hz     & 15020     & 146  &   \\ \hline
Combustor can 2, magnitude in first peak frequency in band 2  & psi    & 15020     & 148  &   \\ \hline
Combustor can 2, magnitude in second peak frequency in band 1 & psi    & 15020     & 152  &   \\ \hline
Combustor can 3, third peak frequency in band 2   & hz     & 15020     & 155  &   \\ \hline
Combustor can 1, magnitude in second peak frequency in band 2 & psi    & 15020     & 157  &   \\ \hline
Combustor can 2, second peak frequency in band 2  & hz     & 15020     & 165  &   \\ \hline
Combustor can 3, first peak frequency in band 1   & hz     & 15020     & 166  &   \\ \hline
Combustor can 4, first peak frequency in band 1   & hz     & 15020     & 167  &   \\ \hline
Combustor can 1, magnitude in first peak frequency in band 1  & psi    & 15020     & 170  &   \\ \hline
Combustor can 4, second peak frequency in band 1  & hz     & 15020     & 172  &   \\ \hline
Combustor can 6, second peak frequency in band 1  & hz     & 15020     & 19   &   \\ \hline
Combustor can 6, magnitude in second peak frequency in band 1 & psi    & 15020     & 53   &   \\ \hline
Combustor can 5, magnitude in second peak frequency in band 1 & psi    & 15020     & 84   &   \\ \hline
Combustor can 5, second peak frequency in band 1  & hz     & 15020     & 139  &   \\ \hline
Combustor can 3, magnitude in third peak frequency in band 1  & psi    & 15020     & 131  &   \\ \hline
Combustor can 3, third peak frequency in band 1   & hz     & 15020     & 160  &   \\ \hline
Combustor can 6, magnitude in third peak frequency in band 1  & psi    & 15020     & 92   &   \\ \hline
Combustor can 6, third peak frequency in band 1   & hz     & 15020     & 128  &   \\ \hline
Combustor can 1, magnitude in third peak frequency in band 1  & psi    & 15020     & 86   &   \\ \hline
Combustor can 1, third peak frequency in band 1   & hz     & 15020     & 161  &   \\ \hline
Combustor can 4, magnitude in third peak frequency in band 1  & psi    & 15020     & 85   &   \\ \hline
Combustor can 4, third peak frequency in band 1   & hz     & 15020     & 122  &   \\ \hline
Combustor can 2, third peak frequency in band 1   & hz     & 15020     & 34   &   \\ \hline
Combustor can 2, magnitude in third peak frequency in band 1  & psi    & 15020     & 124  &   \\ \hline
Combustor can 5, magnitude in third peak frequency in band 1  & psi    & 15020     & 51   &   \\ \hline
Combustor can 5, third peak frequency in band 1   & hz     & 15020     & 56   &   \\ \hline
Center casing, magnitude in first peak frequency in band 2    & psi    & 16226     & 93   &   \\ \hline
Center casing, first peak frequency in band 2     & hz     & 16226     & 164  &   \\ \hline
Center casing, magnitude in second peak frequency in band 2   & psi    & 16226     & 60   &   \\ \hline
Center casing, second peak frequency in band 2    & hz     & 16226     & 142  &   \\ \hline
Center casing, third peak frequency in band 2     & hz     & 16226     & 158  &   \\ \hline
Center casing, magnitude in third peak frequency in band 2    & psi    & 16226     & 173  &   \\ \hline
Center casing, first peak frequency in band 1     & hz     & 16226     & 48   &   \\ \hline
Center casing, second peak frequency in band 1    & hz     & 16226     & 52   &   \\ \hline
Center casing, magnitude in second peak frequency in band 1   & psi    & 16226     & 57   &   \\ \hline
Center casing, magnitude in first peak frequency in band 1    & psi    & 16226     & 103  &   \\ \hline
Center casing, magnitude in third peak frequency in band 1    & psi    & 16226     & 138  &   \\ \hline
Center casing, third peak frequency in band 1     & hz     & 16226     & 169  &   \\ \hline
Combustion chamber exit mass flow     & kg/s   & 17713     & 61   & 17\\ \hline
Lube Oil Pressure    & \degree C    & 18021     & 137  &   \\ \hline
Pressure drop across venturi     & mbar   & 19528     &&   \\ \hline
Center casing, first peak frequency in band 3     & hz     & 20489     &&   \\ \hline
Center casing, second peak frequency in band 3    & hz     & 20489     &&   \\ \hline
Center casing, third peak frequency in band 3     & hz     & 20489     &&   \\ \hline
Center casing, magnitude in first peak frequency in band 3    & psi    & 20489     &&   \\ \hline
Center casing, magnitude in second peak frequency in band 3   & psi    & 20489     &&   \\ \hline
Center casing, magnitude in third peak frequency in band 3    & psi    & 20489     &&   \\ \hline
Turbine interduct pressure & bara & 23497     &&   \\ \hline
\end{longtable}






\end{document}